\documentclass[runningheads]{llncs}
\usepackage[T1]{fontenc}
\usepackage{graphicx}
\usepackage{hyperref}
\usepackage[dvipsnames]{xcolor}
\usepackage{amssymb}
\usepackage{cite}
\usepackage{arydshln}
\usepackage{multirow}
\usepackage{enumitem}
\usepackage{amsmath}    
\usepackage{amssymb}    
\usepackage{mathtools}  
\usepackage{bm}         
\usepackage{booktabs}
\usepackage{float}        
\usepackage{cases}        
\usepackage{booktabs}     
\usepackage{algorithm}
\usepackage{algorithmicx}
\usepackage{algpseudocode}
\usepackage{url}
\usepackage{soul,color}
\algrenewcommand\algorithmicrequire{\textbf{Input:}}
\algrenewcommand\algorithmicensure{\textbf{Output:}}

\begin{document}
%

\title{Analyzing Code Provenance in LLMs for Programming Education}

\titlerunning{Analyzing Code Provenance in LLMs for Programming Education}
%
\author{ Subhankar Maity \inst{1} \and
Faiza Belbachir\inst{1} \and 
Omar Boucelma \inst{2} \and
Guilherme Medeiros Machado \inst{1} 
}
\authorrunning{S. Maity et al.}
%
\institute{LyRIDS, ECE Research Center, Paris, France \and Aix Marseille Univ,  Université de Toulon, CNRS, LIS, Marseille, France \\
\email{\{smaity, fbelbachir, gmedeirosmachado\}@ece.fr},
\email{omar.boucelma@univ-amu.fr}}
\maketitle              
\begin{abstract}

Large language models (LLMs) play an increasing role in programming education, challenging traditional assessment methods as they can generate fully functional code instantly. This disrupts the reliability of student submissions as indicators of learning and effort. To address this issue, code provenance is essential for tracing the origins of code and distinguishing student-authored, AI-generated, and externally sourced components. We extend the Min-K\% probability approach by incorporating synthetic non-member code generation and apply it to Python code for membership inference and provenance tracking. Evaluations cover free LLMs such as GPT-2, DistilGPT-2, and DialoGPT, as well as state-of-the-art commercial models including GPT-5, Gemini 2.5, Mistral, DeepSeek, and LLaMA 3.3. Before evaluating our approach on higher-education student assignments, we validate its feasibility using Stack Overflow data. The results show that free models enable more sensitive code memorization detection than commercial models. By systematically comparing these LLMs, we assess trade-offs between detection accuracy and privacy protection, taking an initial step toward effective assessment of AI-assisted programming education through combined provenance detection and process visibility.


\end{abstract}
\section{Introduction}




The increasing integration of LLMs into education, alongside their use in industry and consumer services, underscores the need for transparency in training data provenance \cite{r5}. In educational contexts, it is essential for instructors to understand which data inform the model outputs. Such understanding builds trust and helps prevent unintended misuse, including biased or unfair
grading. The enormous scale and diversity of datasets fueling LLM development complicate efforts to identify whether specific data samples have contributed to model learning, intensifying debates over intellectual property, data privacy, and ethical AI use \cite{r9}. These challenges are especially acute in education, where programming courses increasingly grapple with sophisticated forms of code plagiarism, often involving AI-generated or online-sourced solutions \cite{r27}. Traditional plagiarism detection mechanisms struggle against such sophisticated methods, jeopardizing academic integrity and legal compliance \cite{r26}.  However, the motivation goes beyond plagiarism detection. We are in a transition period in which LLMs are significantly reshaping how learning is observed and assessed \cite{r5}. Before the advent of generative AI, programming assessments primarily focused on evaluating the final student code for correctness and style, following established summative or formative assessment approaches \cite{r42}. These traditional methods, grounded in product-based evaluation, emphasized the assessment of completed programming assignments rather than the learning journey itself. Coding exercises reliably reflected students' understanding, effort, and problem-solving skills \cite{r23}. However, there is now a growing need to shift towards assessing the learning process itself, how students develop, modify, and integrate code over time. As LLMs can generate fully functional solutions instantly, code provenance, which tracks the origin and evolution of code segments, becomes essential for gaining deeper insight into student learning and supporting authentic process-oriented assessment. 



This shift demands a fundamental change in assessment practices, from focusing solely on final product correctness to emphasizing the student’s learning process. New formative assessment modalities are needed to make student thinking visible and support scalable instructor workflows. Code provenance technique offers a transformative solution by providing an overview of AI usage patterns across classes, acting as a pedagogical filter \cite{r28}. This enables instructors to prioritize formative feedback, identify students relying heavily on AI, or pinpoint widespread conceptual outsourcing, facilitating targeted interventions previously infeasible at scale. For example, an instructor could quickly identify students who rely almost exclusively on AI-generated code or pinpoint a specific programming concept (e.g., \textit{data structures}, \textit{recursion}, etc.) that a large portion of the class is outsourcing to AI. This enables a targeted intervention and makes the widespread adoption of formative assessment practices feasible for the first time. In this way, code provenance is not just an integrity tool; it serves as gate for a more effective and scalable homework assessment in the age of generative AI. Despite advances in membership inference for text data such as Wikipedia and copyright-protected books \cite{r1}, research specifically addressing code remains sparse. Code data presents unique challenges compared to natural language text due to its structured syntax, dynamic behavior, and legal sensitivities \cite{r17}. These characteristics make provenance analysis especially crucial given the growing role of code generation in LLMs. Unlike plain text, code relies on precise formatting, hierarchical structure, and functional dependencies, which complicate traditional membership inference methods. To address this gap, we present the first comprehensive study applying membership inference attacks not only for code plagiarism detection but also for establishing a practical framework for scalable, process-oriented assessment methods. By systematically comparing free and commercial LLMs using the proposed method, this research takes an initial step toward assessing the trade-offs between detection accuracy and privacy protection in programming education. The central research question guiding this study is: \textit{To what extent can membership inference attacks on LLMs effectively detect code provenance to enable pedagogically meaningful, process-oriented assessment in AI-assisted programming education?} This investigation aims to inform ongoing efforts to enhance assessment practices through a proof-of-concept in AI-assisted programming education. Specifically, this study makes the following contributions.

\begin{itemize}
    \item We extend the Min-K\% probability approach by introducing a method for generating synthetic non-member code, specifically tailored to Python code data for provenance tracing of LLMs. These adaptations account for the unique structural and semantic characteristics of code data in educational settings, ensuring the method’s applicability beyond text to programming code. Furthermore, the approach can be readily extended to other programming languages.
    

    \item We evaluate multiple free models, including variants of GPT-2, Distil GPT-2, and DialoGPT, alongside state-of-the-art commercial models such as GPT-5, Gemini 2.5, Mistral, DeepSeek, and LLaMA 3.3 using the proposed methodology.

    \item To verify the applicability of this method in real-world educational settings, we conduct a proof‑of‑concept experiment on Python programming assignments from higher‑education students. This experiment evaluates the extent of memorization in general‑purpose LLMs, as these models are most commonly used by students for programming tasks.


    \item We fill a critical gap in the literature by analyzing the trade-offs between privacy protection and detection effectiveness that educational institutions face when choosing between free and paid AI models.
\end{itemize}

\section{Related Work}


\noindent \textbf{Membership Inference Attacks (MIAs):} The primary objective of MIAs is to determine whether an arbitrary data sample was part of the training dataset of a target model \cite{r44}. Model memorization occurs when LLMs reproduce segments of their training data verbatim or nearly so, raising concerns about unintended disclosure of sensitive or proprietary information \cite{r45}. Data provenance, defined as the traceability of the origin and history of data modification, provides an essential lens to ensure accountability in AI systems and educational applications \cite{r46}. Previous research has focused largely on textual data such as Wikipedia and copyrighted book collections \cite{r1}, while code data, with its inherent structural and semantic complexity and its unique importance in educational contexts, has received comparatively less attention. MIAs have evolved from basic nearest-neighbor detection to sophisticated probabilistic metrics, including the Min-K\% approach, computing the average log-probability of the least likely tokens in given samples. \cite{r3} and \cite{r4} have emphasized the privacy risks and legal implications of data leakage in LLMs, sparking debate about responsible AI in practice. However, these works largely overlook code data, which is more sensitive due to its embodiment of intellectual property, unique problem-solving logic, and potential exposure of confidential business or security processes.


\noindent \textbf{Traditional Plagiarism Detection in Programming Education:} Plagiarism detection in programming education has traditionally relied on analyzing code similarity, employing static metrics, and leveraging stylometric techniques \cite{r18, r66}. Notable approaches include structural comparison methods, such as those employed in MOSS and JPlag, behavioral analysis systems, like BPlag, which measures program execution similarity, and consensus-based frameworks that combine multiple detection tools to improve accuracy \cite{r60, r67}. Recent work has also explored machine learning approaches \cite{r61}, with SVM-based systems showing promise for detecting plagiarism of Python code in undergraduate settings \cite{r62}, and visual proof systems that provide transparent evidence of code similarity for educators \cite{r59}. 

\noindent \textbf{Limitations of Existing Approaches with AI-Generated Code:} However, these traditional approaches face significant limitations in the educational landscape of today. Although they effectively identify similar code structures or patterns, they cannot distinguish whether these similarities arise from exposure to model training data or from direct AI generation \cite{r27, r18}, a crucial distinction for an accurate educational evaluation. This limitation becomes increasingly problematic as students nowadays engage more in general-purpose LLMs, making it essential for educators to understand not only whether code is AI-generated, but also the source and nature of that generation \cite{r28}.

\section{Task Definition}

The membership inference attack problem can be formally defined as follows. Let \( f_{\theta} \) be a language model trained on a dataset \( D = \{ z_i \}_{i=1}^n \) sampled from an underlying distribution. The objective is to learn a detector \( h \) that infers the membership status of any arbitrary data point \( x \), formally expressed as
\[
h(x, f_{\theta}) \to \{0, 1\},
\]
where `0' denotes non-membership and `1' indicates membership. We assume a black-box setting in which the detector can only query \( f_{\theta} \) to obtain token probabilities for the input \( x \). Given a target language model \( M \) equivalent to \( f_{\theta} \), trained on a code dataset \( D \), the task is to determine whether a query code sample \( x \) is part of \( D \). This problem is posed as binary classification:
\[
y =
\begin{cases}
1 & \text{if } x \in D \quad \text{(member)} \\
0 & \text{if } x \notin D \quad \text{(non-member)}.
\end{cases}
\]
\textit{Member} data \( D \) consists of code samples encountered during training, for which the model is expected to assign higher confidence (lower perplexity). Conversely, \textit{non-member} data \( D' \) comprises unseen code samples eliciting lower confidence (higher perplexity). In practical terms, such a membership detector can be employed to analyze coding assignments and determine if a given code snippet was likely seen during model training. A positive membership classification indicates that the model has ‘seen’ or memorized the code, which has important implications for maintaining assessment fairness and supporting authentic student learning.


\section{Methodology}




At the core of our pipeline is the extended Min-K\% membership algorithm \cite{r1}, implemented for code snippets in a modular fashion to enable robust and comparative evaluation across both free and commercial LLMs (Algorithm \ref{alg:min-k-python-fixed-k}). As a novel contribution, we introduce synthetic non-member code generation to simulate code samples absent from a given model’s training data (Algorithm \ref{alg:synthetic-code-generation}). Although this study focuses on Python code, the approach is designed to be adaptable to other programming languages and educational contexts.

\noindent \textbf{The Min-K\% Algorithm:} The core intuition behind the algorithm is that language models assign differential confidence levels to samples they have seen during training versus unseen samples. Training data typically yield higher token probabilities (and thus lower perplexities). Focusing on the lowest \( K\% \) of token probabilities highlights the model's most uncertain predictions, which are more discriminative than average perplexity, enhancing detection accuracy \cite{r1}. Algorithm \ref{alg:min-k-python-fixed-k} details the membership inference detection procedure for code snippets.

\begin{algorithm}[H]
\caption{Membership Inference Detection for \textit{Code Snippets}}
\label{alg:min-k-python-fixed-k}

\begin{algorithmic}[1]

  \Require Python code token sequence \( x = x_1, x_2, \ldots, x_n \); threshold \(\tau\); parameter \(K\%\)
  \Ensure Membership classification (Member/Non-member)
  \For{\(i = 1\) to \(n\)}
      \State Compute log probability \(p(x_i | x_1,\ldots, x_{i-1})\) via model \(f_\theta\)
  \EndFor
  \State Sort log probabilities ascending \(\to L_{\text{sorted}}(x)\)
  
  \State Extract the lowest \( k \) log-probabilities: \( \text{Min-}k(x) = L_{\text{sorted}}(x)[1 : k] \)
  \State Compute score: $S_{\text{MinK}}(x) = \frac{1}{k}\sum_{i=1}^k \text{Min-k\%}(x)[i]$
  \If { \( S_{\text{MinK}}(x) \geq \tau \) }
    \State \Return \texttt{Non-Member} \Comment{Code snippet likely unseen}
  \Else
    \State \Return \texttt{Member} \Comment{Code snippet likely in training data}
  \EndIf
\end{algorithmic}
\end{algorithm}

\begin{figure}[ht]
    \centering
    \includegraphics[width=0.88\textwidth]{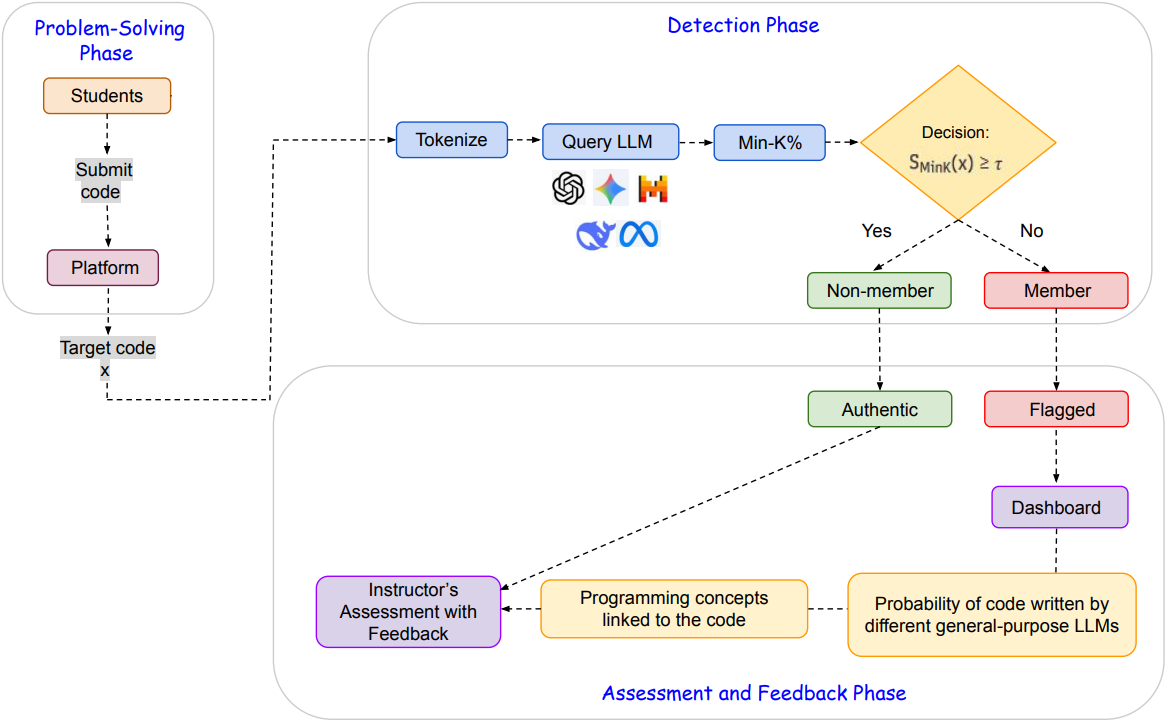} 
    \caption{Workflow for detecting LLM-generated code in students' code using the Min-K\% membership inference attack. 
    }

    \label{fig:mink}
\end{figure}

 \noindent \textbf{Workflow Phases:} Fig. \ref{fig:mink} presents the workflow for detecting LLM-generated code in coding homework using Min-K\% Membership Inference Attack. Each phase of the workflow is as follows.

 \begin{enumerate}[label=\alph*.]
     
 \item \textbf{Problem-solving Phase:} The workflow begins with students submitting source code to a Learning Management System (LMS). A target code sample $x$ is selected for analysis. 
 
 \item \textbf{Detection Phase:} In the Detection Phase, $x$ is tokenized and fed to the target LLM $f_\theta$. The Min-K\% algorithm (Algorithm \ref{alg:min-k-python-fixed-k}) processes the token probabilities: it calculates and sorts the log-probabilities, then computes the average \( S_{\text{MinK}}(x) \) of the lowest $K\%$ values. The score is compared against a pre-defined threshold ($\tau$) to classify the sample as a \textit{Member} (flagged) or a \textit{Non-member} (Authentic). 
 
 \item \textbf{Assessment and Feedback Phase:} Finally, in this phase, the membership result is presented to the instructor through a dashboard where they can check the probability that the flagged code was generated by different general-purpose LLMs. An instructor can also analyze the specific programming concepts related to the flagged code, gaining insights into which concepts students use LLMs most frequently and understanding their problem-solving processes. This enables educators to provide targeted feedback and assessments tailored to each student’s usage of AI-generated code.

 \end{enumerate}
 
\noindent \textbf{Threshold Selection:} In our methodology, percentile-based thresholding is used to convert continuous membership scores into binary membership predictions. Following \cite{r68}, thresholds (\(\tau\)) are calculated from the empirical distributions of membership scores for member and non-member samples. The average of the 75\textsuperscript{th} percentile of member scores and the 25\textsuperscript{th} percentile of non-member scores is chosen as the threshold. This choice is prudent as it effectively separates the two distributions by capturing a central tendency while minimizing the impact of outliers and noise. It ensures more stable and interpretable decision boundaries compared to heuristic or midpoint methods. Such robustness is crucial given the variability in code samples and complex behavior of language models. 


\section{Experiments} \label{non}

In this section, we describe the datasets used, detailing the member and synthetic non-member data, outline the external feasibility check conducted on Stack Overflow, present the experimental setup, and discuss the metrics employed to evaluate membership inference performance.\\ \\
\noindent \textbf{Member Data:}
Python code samples are extracted from the ProCQA dataset \cite{r20}. ProCQA contains approximately 1.008 million records for Python, drawn from a collection spanning 11 popular programming languages, including Python, Java, JavaScript, Ruby, C, C++, C\#, Rust, PHP, Lisp, and Go,  sourced from Stack Overflow. Each record in ProCQA consists of a `\textit{question}', `\textit{description}', and `\textit{answer}'. For our study, only the `\textit{answer}' record was used. These samples serve as our \textit{member} code corpus for subsequent membership inference attacks and provenance tracking. The focus was narrowed to Python, which is the most popular language on Stack Overflow, widely used in educational settings, and the language of our curated student programming assignments. Table \ref{tab:python_concepts} presents the key Python programming concepts and the number of related programming questions in the ProCQA dataset. The set of programming concepts was adopted from the core topics\footnote{\href{https://ocw.mit.edu/courses/6-0001-introduction-to-computer-science-and-programming-in-python-fall-2016/pages/readings/}{MIT OpenCourseWare}} and the recommended textbook on MIT OpenCourseWare.

\begin{table}[htbp]
\small
\centering
\caption{Distribution of key Python programming concepts and the number of programming questions related to each concept in the ProCQA dataset.}
\label{tab:python_concepts}
\begin{tabular}{|l|r|}
\hline
\textbf{Programming Concept} & \textbf{Frequency} \\
\hline
Tuples, Lists, Aliasing, Mutability, Cloning & $\sim$1.5M \\
String manipulation, Guess and Check, Approximations, Bisection & $\sim$1.26M \\
Decomposition, Abstractions, Functions & $\sim$993K \\
Branching and iterations & $\sim$963K \\
Recursion, Dictionaries & $\sim$806K \\
Testing, Debugging, Exceptions, Assertions & $\sim$596K \\
Data structures and algorithms & $\sim$569K \\
\hline
\end{tabular}
\end{table}

\noindent \textbf{Synthetic Non‑Member Data:}
Inspired by prior work on data augmentation and code transformation for evaluating memorization in language models \cite{r48, r50, r52, r54, r56}, we introduce a novel method (Algorithm \ref{alg:synthetic-code-generation}) to synthetically generate non-member code. This technique provides a controlled and reproducible way to simulate non‑member samples, code segments that are algorithmically derived from, yet distinct from, the original training‑like code. We generate diverse synthetic variations of the member code using five weighted randomization transformation strategies, designed to alter the surface form while preserving, partially preserving, or intentionally perturbing semantic intent. The transformation strategies are as follows:

\begin{enumerate}
    
\item Semantic Transformation (15\%): Randomizes the name of variables and functions while modifying, removing, or paraphrasing comments to prevent memorization based on identifiers and text from natural language \cite{r48}.

\item Structural Transformation (15\%): Alters indentation, spacing, formatting, and reorders control flow and logical blocks to disrupt superficial visual and logical patterns \cite{r50}.

\item Aggressive Transformation (40\%): Injects redundant or equivalent logic and deliberately breaks code logic while maintaining syntactic correctness, challenging model robustness against subtle memorization \cite{r52}.

\item Cross Language Conversion (25\%): Rewrites code to adopt the syntax and idioms of diverse languages and frameworks, e.g., JavaScript, Java, C++, GraphQL, Jupyter notebooks, adding syntactic and structural variety without altering its semantic intent \cite{r54}.

\item Novel Pattern Generation (5\%): Generates entirely novel code structures beyond Python, including SQL/NoSQL queries, orchestration patterns (Kubernetes, Docker), and configuration scripts (Terraform), offering linguistically distant and structurally diverse examples \cite{r56}.

\end{enumerate}

\begin{algorithm}[H]
\caption{Synthetic Non-Member Code Generation for Code Data}
\label{alg:synthetic-code-generation}

\begin{algorithmic}[1]

  \Require Python code snippets extracted from the ProCQA dataset \( C \)
  \Ensure Synthetic non-member code set \( C' \)
  \State Initialize \( C' \gets \varnothing \)
  \For{each code sample \( c \in C \)}
      \State Randomly draw \( r \sim \text{Uniform}(0, 1) \)
      \If{ \( r < 0.15 \) }
          \State \( c' \gets \textsc{SemanticTransformation}(c) \) 
      \ElsIf{ \( r < 0.30 \) }
          \State \( c' \gets \textsc{StructuralTransformation}(c) \)
        
      \ElsIf{ \( r < 0.70 \) }
          \State \( c' \gets \textsc{AggressiveTransformation}(c) \)
          
      \ElsIf{ \( r < 0.95 \) }
          \State \( c' \gets \textsc{CrossLanguageConversion}(c) \)
      \Else
          \State \( c' \gets \textsc{NovelPatternGeneration}(c) \)
      \EndIf
      \State Add \( c' \) to \( C' \)
  \EndFor
  \State \Return \( C' \)
\end{algorithmic}
\end{algorithm}


To validate that the synthetic non‑member samples are distinct from the member data, we compute abstract syntax tree (AST)‑based distance \cite{r71} by first parsing each code snippet into an AST and then measuring the normalized tree‑edit distance between the original member AST and its synthetic counterpart over 500 code pairs. A distance of 0.996 (mean) indicates that synthesizing the transformed code from the original AST requires extensive structural edits, showing that the synthetic instances are structurally and syntactically far from the original data.\\

\noindent \textbf{External Feasibility Check on Stack Overflow:} Before evaluating our approach on student programming assignments, we assess its feasibility using public Stack Overflow data, as represented in the ProCQA dataset. Stack Overflow, widely used by students for coding assistance, provides extensive community‑curated programming knowledge and practical problem‑solving insights \cite{r40}. By first evaluating the method on code derived from Stack Overflow code snippets and then on student assignments (Section \ref{casestudy}), we ensure that our approach generalizes beyond a single in‑house code base and remains applicable in realistic, education‑relevant settings.\\



\noindent \textbf{Experimental Setup:} Our experimental setup provides an evaluation framework across multiple LLMs. Data are split into 30\% for threshold calibration and 70\% for evaluation and results analysis. Following \cite{r1}, we explored different fixed values of the hyperparameter \( k \), such as \( k = 5, 10, 20 \), to identify the optimal number of lowest token probabilities to consider when computing the Min-K score. We found that \( k = 10 \) works best for membership inference detection. We evaluate both free models, including GPT-2, Distil GPT-2, and DialoGPT variants, on approximately 3000 samples per class via the Hugging Face platform. Furthermore, we evaluated commercial API-based models, including GPT-5, Gemini-2.5-Flash, Mistral-Small-24B-Instruct, DeepSeek-V3, and LLaMA-3.3-70B-Instruct, on approximately 200 samples per class due to budget constraints, using OpenRouter APIs\footnote{\url{https://openrouter.ai/models}}. We focus on general-purpose models because these models are the ones most commonly accessible to students and educators in real-world educational settings \cite{r69}. Moreover, as of October 2025, recent coding LLM leaderboard\footnote{\href{https://www.vellum.ai/best-llm-for-coding?utm_source=google&utm_medium=organic}{Coding LLM Leaderboard}} highlight that general-purpose models within the GPT, DeepSeek, Llama, Gemini, and Mistral families consistently rank among the top-performing systems. We ran all our experiments on NVIDIA Tesla T4 GPU. The code and dataset are available at \href{https://anonymous.4open.science/r/student-code-provenance-LLMs-MinK-FF0B/}{Anonymized GitHub Repository}.\\

\noindent \textbf{Metrics:}
Following prior work \cite{r1}, we calculate the detection effectiveness using the true positive rate (TPR) and the false positive rate (FPR). As core metrics, we report the area under the ROC curve (AUC), indicating whether the method assigns higher scores to training samples than to non-training ones, and TPR@5\%FPR, which measures the proportion of real training samples correctly detected with only 5\% false positives. In addition, we also report precision and F1 scores to provide a more comprehensive assessment of performance.

\section{Results and Analysis}

\begin{figure}[ht]
    \centering
    \includegraphics[width=0.65\textwidth]{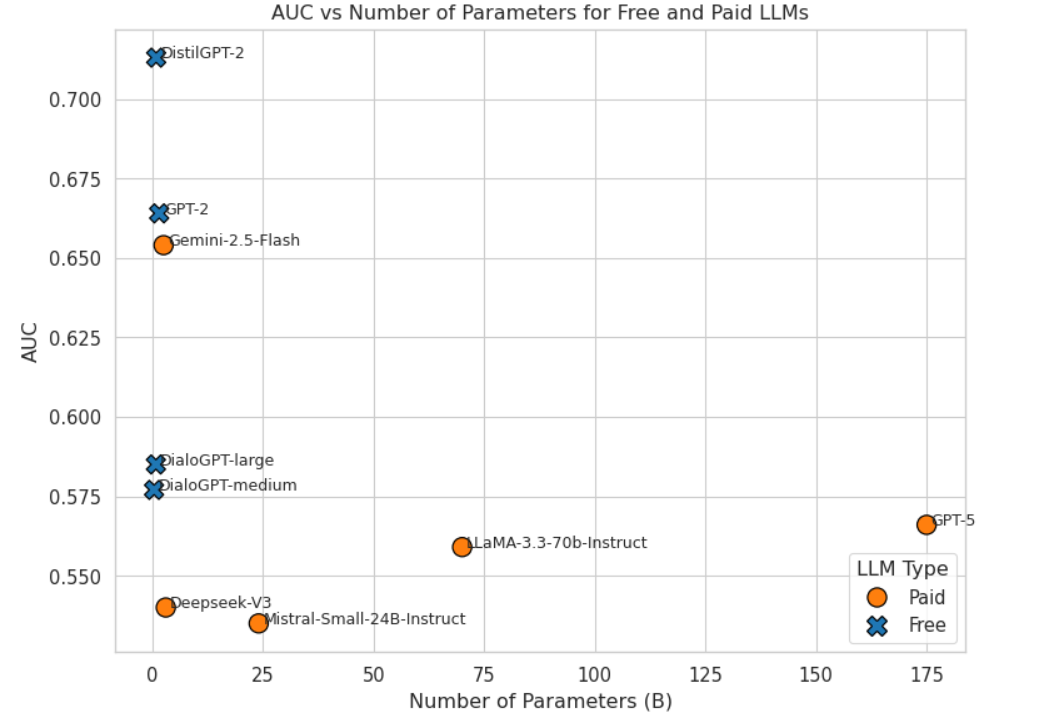} 
    \caption{We show AUC performance across various LLMs plotted against their model size (number of parameters, in billions (B)). The models are grouped into free and paid categories to illustrate differences in membership inference detection effectiveness relative to model scale.}
    \label{fig:aucvsparam}
\end{figure}

The results on general-purpose LLMs and Stack Overflow code are first presented, which serve as a technical validation before presenting the educational proof-of-concept validation. The membership inference detection performance of the free LLMs demonstrates notable variation across the evaluated metrics. As shown in Table \ref{tab:llm-metrics}, DistilGPT-2 emerges as the top performer among the free models, achieving the highest AUC of 0.713, along with leading precision (0.680) and F1 score (0.674). This means DistilGPT-2 demonstrates superior capability in distinguishing member from non-member data and detecting code provenance among free LLMs. GPT-2 achieves the strongest TPR@5\%FPR at 18.19\%, indicating its superior capability to identify member samples while minimizing false alarms. In contrast, the DialoGPT models, both medium and large versions, show relatively lower performance, pointing to the influence of architectural and training differences on the detectability of membership inference.
Paid LLMs generally report lower membership inference detection scores. As presented in Table \ref{tab:llm-metrics}, Gemini-2.5-Flash stands out with the highest AUC of 0.654, reflecting its superior overall membership inference detection capability relative to its commercial peers. GPT-5 outperforms other paid LLMs in terms of TPR@5\%FPR, achieving the highest value of 10.14\%, as well as precision with a score of 0.569. Notably, LLaMA-3.3-70B-Instruct exhibits the strongest performance in F1 score (0.667), indicating a balanced trade-off between precision and recall, and thus robust membership inference detection overall. Importantly, these results show that some smaller or medium-sized free LLMs can outperform larger paid counterparts in membership inference detection, underscoring that model size alone does not determine leakage potential. Instead, detection capability depends on a complex interplay of model architecture, training data, and implementation specifics. Fig. \ref{fig:aucvsparam} visually presents AUC performance relative to model size (in billions of parameters) for both free and paid LLM categories. The figure indicates no straightforward correlation between model size and inference accuracy. Some smaller or medium-sized free LLMs outperform larger paid counterparts, underscoring that model size alone does not determine leakage potential.

\begin{table}[h]
\centering
\caption{Performance metrics of general-purpose LLMs on membership inference evaluation. The top portion presents \textit{free} LLMs, while the bottom portion lists \textit{paid/commercial} LLMs. The highest scores within each group are highlighted in \textbf{bold}.}
\label{tab:llm-metrics}
\begin{tabular}{|l|c|c|c|c|}
\hline
\textbf{Model} & \textbf{AUC} & \textbf{TPR@5\%FPR} & \textbf{Precision} & \textbf{F1} \\
\hline
DialoGPT-medium        & 0.577 & 9.24  & 0.564 & 0.543 \\
DialoGPT-large         & 0.585 & 10.28 & 0.608 & 0.471 \\
GPT-2                  & 0.664 & \textbf{18.19} & 0.651 & 0.631 \\
DistilGPT-2            & \textbf{0.713} & 17.24 & \textbf{0.680} & \textbf{0.674} \\
\hline\hline
Gemini-2.5-Flash       & \textbf{0.654} & 6.16  & 0.567 & 0.666 \\
Mistral-Small-24B-Instruct & 0.535 & 7.29  & 0.509 & 0.639 \\
DeepSeek-V3            & 0.540 & 5.86  & 0.500 & 0.666 \\
LLaMA-3.3-70B-Instruct & 0.559 & 7.48  & 0.557 & \textbf{0.667} \\
GPT-5                  & 0.566 & \textbf{10.14} & \textbf{0.569} & 0.577 \\
\hline
\end{tabular}
\end{table}


As seen from Table \ref{tab:llm-metrics}, paid LLMs generally show lower membership inference detection performance compared to free models, primarily due to intentional design choices made by providers aimed at enhancing user privacy and data protection. Constraints such as limited inference samples arise from the higher computational costs and usage fees associated with paid services, which restrict extensive membership testing. In addition, commercial providers integrate advanced privacy-preserving mechanisms \cite{r35}. These include techniques such as differential privacy and noise injection, which effectively reduce memorization and protect against membership inference attacks \cite{r35}. Architectural innovations focusing on security, combined with API restrictions that limit data exposure, further diminish the attack surface available to adversaries. Training techniques emphasizing regularization and selective data exposure also play a crucial role in mitigating membership leakage \cite{r35}. Furthermore, differences in the way token probabilities are computed and aggregated influence the success of inference detection. These factors collectively highlight the trade-off between maximizing detection accuracy and maintaining privacy safeguards. The findings underscore the need to select models carefully according to the privacy and security requirements of the deployment context. Free models typically enable more effective detection of membership inference risks, while paid LLMs exhibit differing behaviors in memorization that may affect detection sensitivity. Moreover, many free LLMs evaluated are earlier or smaller-scale versions of paid models. Although these models often facilitate stronger detection signals, it remains essential to assess their broader capabilities, including dimensions beyond code generation quality. 


\section{Proof‑of‑Concept Evaluation on Student Python Programming Assignments}\label{casestudy}

\noindent \textbf{Experiments:} To validate the applicability of our approach in authentic educational contexts, we conducted a proof-of-concept on Python programming assignments collected from a data science hackathon involving higher education students. In this hackathon, student groups were tasked with developing the best-performing classification model for the well-known Titanic dataset, predicting passenger survival outcomes. Each group consisted of seven students and submitted one integrated Jupyter notebook as their final solution. For this study, we selected submissions from five groups who provided explicit consent for the use of their code. In addition, these groups voluntarily shared their interaction logs with LLMs used during the hackathon for code generation and completion. These conversation records served as the ground truth for our provenance analysis. We applied the enhanced Min-K\% probability method to these code submissions to estimate the likelihood that different code segments were generated by general-purpose LLMs. The analysis focused on identifying where in the development process, such as package imports, data loading, feature engineering, model training, or evaluation, students relied on LLM assistance. From the shared interaction logs, we extracted both code and associated comments to establish reference annotations of LLM-generated content. To align Min-K\% predictions with the ground truth from chat logs, we applied four complementary matching strategies: exact matching after normalization, character-level fuzzy matching based on sequence similarity, token-level fuzzy matching using code token sequences, and set-based token matching via Jaccard similarity. Comparing these annotations with the Min-K\%-based predictions allowed us to evaluate the accuracy of provenance detection. This proof-of-concept thus demonstrates the method’s practical viability in real classroom-like settings and its potential to reveal fine-grained insights into how students integrate AI assistance into their coding workflows.

\noindent \textbf{Results:} Table \ref{stud} summarizes the provenance detection performance of the enhanced Min-K\% method across the five analyzed student groups. Each group represents an independent submission from the hackathon dataset. 

\begin{table}[h] \centering \caption{Performance of the enhanced Min-K\% method on student group submissions.} 
\label{stud}
\begin{tabular}{|c|c|c|c|c|} 
\hline 
\textbf{Group} & \textbf{AUC} & \textbf{TPR@5\%FPR} & \textbf{Precision} & \textbf{F1} \\ \hline 
Group 1 & 0.584 & 6.25 & 0.723 & 0.780 \\ 
Group 2 & 0.594 & 7.03 & 0.895 & 0.757 \\ 
Group 3 & 0.640 & 8.24 & 0.688 & 0.700 \\ 
Group 4 & 0.644 & 7.86 & 0.542 & 0.644 \\ 
Group 5 & 0.629 & 7.10 & 0.233 & 0.368 \\ \hline 
\end{tabular} 
\end{table}

Overall, the AUC values (0.58–0.64) indicate moderate discriminative ability in distinguishing LLM-generated from student-authored code segments. The TPR@5\%FPR values remain below 9\%, reflecting the challenge of achieving strong detection sensitivity at very low false positive rates in authentic, line-level code data. Precision and F1 scores show notable variation across groups, suggesting different levels of reliance on and post-editing of AI-generated code. Groups 1 and 2 achieved higher precision (0.72–0.89) and balanced F1 scores, indicating clearer separability between human-written and LLM-derived content. Beyond quantitative metrics, a qualitative inspection of matched segments revealed that most groups used LLM-generated content primarily for feature engineering tasks, followed by assistance in importing packages, training models, and generating evaluation plots. These patterns provided instructors with fine-grained insights into where students sought AI assistance, enabling more targeted feedback on specific coding stages and supporting process-oriented assessment. Additionally, when estimating the probability of code written by different general-purpose LLMs, our implementation consistently attributed the highest likelihood to GPT-5 (ChatGPT) across all groups. This observation was further confirmed by students’ shared conversation logs, which verified that ChatGPT was indeed the primary model used throughout their coding process. 



\section{Conclusion}

This study extends the Min-K\% probability approach for tracing code provenance in programming education, enabling differentiation between student-authored and LLM-generated code. Through comprehensive evaluations on both open and commercial models and a real-world proof-of-concept with student hackathon submissions, the results demonstrate that the method can reliably identify patterns of AI assistance at a fine-grained level. Free models showed greater sensitivity for detecting memorization, while commercial models provided stronger privacy defenses, underscoring a key trade-off for educational deployments. The proof-of-concept further revealed that students primarily used ChatGPT (GPT-5) for feature engineering, model training, and result visualization tasks. We acknowledge that our method detects training data membership, potentially conflating AI generation with manual retrieval from web sources like Stack Overflow. While these actions involve distinct cognitive processes—searching versus prompting—both represent a reliance on external solutions rather than independent synthesis. By identifying such non-original code, this work offers a foundational step toward process-oriented assessment, moving beyond final-product evaluation while future extensions aim to disentangle specific provenance. These insights highlight the dual value of provenance analysis for both privacy evaluation and pedagogical feedback, offering educators a practical framework to promote transparency, integrity, and process-oriented assessment in AI-assisted programming education.

\bibliographystyle{splncs04}
\bibliography{mybib}

\end{document}